\documentclass[letterpaper, 10 pt, conference]{ieeeconf}
\IEEEoverridecommandlockouts

\usepackage{graphicx}
\usepackage{booktabs}
\usepackage{rotating}
\usepackage{stfloats}
\usepackage{amsmath}
\usepackage{amssymb}      
\usepackage{url}

\title{\LARGE \bf
Learning Faster without Deeper Networks: A*-Inspired Batch Selection for Efficient CNN Training
}

\author{
Anxhelo Shehu$^{1}$,
Enes Stastoli$^{1}$,
Arben Cela$^{1,2}$%
\thanks{$^{1}$AI Laboratory, University Metropolitan Tirana, Tirana, Albania.}%
\thanks{$^{2}$Laboratory of Images Signals and Intelligent Systems, ESIEE Paris, Paris, France.}%
\thanks{Emails: \{anxheloshehu, estastoli\}@umt.edu.al, arben.celaa@gmail.com}%
}

\begin{document}
\maketitle
\thispagestyle{empty}
\pagestyle{empty}

\begin{abstract}
Common practice when training Convolutional Neural Networks (CNNs) is to utilize randomly shuffled mini-batches. During this stage of the training process, two limitations are faced. Firstly the slowing down of convergence, and secondly the limited learning signal contribution due to large quantities of samples that are considered easy while training CNNs.

In our work, we aim to address these inefficiencies by proposing \emph{A*-Inspired Batch Selection} (A*-BS). This is a lightweight and model-agnostic training strategy that formulates mini-batch scheduling as a heuristic search problem. In this type of batch selection process each batch is treated as a node in a search space and ranked using an A*-like score by combining a loss-based difficulty measure with a reuse penalty. The two main benefits of this method are: informative gradient updates encouragement and batch diversity selection throughout the training process. The proposed method does not modify network architectures or optimization algorithms and can be seamlessly integrated into existing training pipelines. We evaluate the approach on the twelve 2D classification tasks of the MedMNIST-v2 benchmark, using a deliberately simple architecture of approximately $2.25 \times 10^{5}$ parameters and comparing against the ResNet-18 and ResNet-50 baselines reported by the benchmark.

On half of these tasks, the lightweight model combined with A*-BS reaches higher accuracy and AUC than both ResNet baselines, with relative gains of up to 15\%. Moreover, an ablation under identical CNN architecture and hyper-parameters shows that A*-BS outperforms random batch shuffling on all twelve MedMNIST tasks. Wall-clock measurements further show that the lightweight CNN with A*-BS trains substantially faster than ResNet-18 and ResNet-50 on identical hardware.

Indicative in these results is that intelligent batch ordering can partially compensate for reduced architectural complexity, thus offering a computationally efficient alternative to deeper models. Furthermore, the reliability of the proposed strategy is reinforced by its strong performance with a simple CNN, even when compared to deeper and more sophisticated architectures.
\end{abstract}

\vspace{0.5em}
\noindent\textbf{\textit{Index Terms---}}\textit{
Batch Selection, Heuristic Search, A* algorithm, Curriculum Learning, CNN, MedMNIST-v2, Ablation Study}
\vspace{0.5em}
\section{INTRODUCTION}

Deep learning models, and in particular CNNs, have become the dominant approach for image classification tasks across a wide range of domains, including medical imaging, autonomous systems, and computer vision. Despite their success, CNNs typically require extensive training on large datasets and long optimization times, which remain significant challenges, especially under limited computational resources~\cite{alzubaidi2021review}.
By using randomly shuffled mini-batches and stochastic gradient descent in training pipelines, it is implicitly assumed that all of the training samples will contribute equally to the learning process.

In practice, however, things differ. Usually, during the early stages of training, a large number of \emph{easy} samples from the dataset is classified correctly with the downside of providing limited informative gradients in the later stages. The motivation for research into curriculum learning and adaptive data selection strategies stemmed from the exact issue of frequent wasted computational effort on samples that contribute little to performance improvement, whereas more challenging or informative samples are usually underutilized. The aim of this research is to present training samples in a more meaningful selection order to accelerate convergence and improve generalization.

Inspired by heuristic search and pathfinding algorithms, this work proposes a novel batch scheduling strategy motivated by the A* algorithm~\cite{astar}. Instead of selecting mini-batches randomly, we treat each batch as a node in a search space and prioritize training based on an A*-like evaluation function that balances batch difficulty and exploration. The A* algorithm is widely known for efficiently navigating search spaces by combining heuristic estimates with accumulated costs, and its theoretical properties have been extensively studied in the literature~\cite{lv_astar}. Through the incorporation of these principles into the training process, and without requiring changes to the network architecture or the optimization algorithms we aim to guide the optimizer toward more informative regions of the data space.

As a result, the proposed approach can be easily integrated into existing CNN training pipelines because it is lightweight and applicable across different architectures. It is particularly well suited for low-resolution medical imaging datasets, where increasing model depth may not be justified and efficient use of training data is crucial. All programs, experimental protocols, and implementation details are made publicly available to ensure reproducibility and facilitate further research\footnote{\url{https://github.com/AI-Lab-UMT/astar-batch-selection-CNN}}.

The contributions of this paper are: a controlled ablation showing that A*-BS outperforms random batch shuffling across all twelve MedMNIST 2D tasks under identical CNN architecture and training settings, and a wall-clock comparison showing that the lightweight CNN with A*-BS trains substantially faster than ResNet-18/-50 on the same hardware. The boundaries of the comparison with the ResNet results from~\cite{medmnistv2} are explicitly discussed in Sec.~IV-A.

\section{RELATED WORK}

In recent years, a significant amount of research has been conducted to improve the optimization and training efficiency of deep learning models, particularly Convolutional Neural Networks (CNNs), as training large-scale models remains computationally expensive and challenging. A significant body of work has focused on curriculum learning and self-paced learning strategies, where training samples are presented in a meaningful order to improve convergence and stability, as introduced by Bengio et al.~\cite{bengio2009curriculum} and later extended by Kumar et al.~\cite{kumar2010self}.

Closely related approaches include hard example mining and loss-aware learning, where difficult samples are emphasized during training, such as Online Hard Example Mining~\cite{shrivastava2016ohem} and Focal Loss~\cite{lin2017focal}, which aim to mitigate class imbalance and improve robustness. It has been demonstrated in the work of ~\cite{loshchilov2016online, katharopoulos2018importance}, through importance sampling and selective training strategies, that not all samples contribute equally to gradient updates and overall model improvement. This point, in more recent loss-aware sampling approaches, it has been further refined by reshaping the sampling distribution in loss statistics. As an example, the Gaussian loss-weighted sampling~\cite{gaussian_sampler} can be taken, where improvements in training efficiency and robustness are observed. The cause for such advances can be attributed to the probabilistic favor of informative samples and the reduction of the influence of both trivial and extreme-loss instances. The discussed sampling-based strategies primarily focus on adjusting the selection of individual samples during the training phase.

Building upon this observation, more recent works have explored dynamic data selection, pruning, and prioritized training mechanisms. Empirical analyses reveal that certain samples consistently have a stronger impact on generalization, while others may become redundant as training progresses~\cite{toneva2019empirical}. Consequently, principled data pruning and prioritized learning frameworks have been proposed~\cite{sorscher2022beyond, mindermann2022prioritized, infobatch}, showing that intelligent data selection can reduce computational cost without sacrificing performance. In a similar direction, coreset-based and data-efficient training frameworks, such as DeepCore~\cite{guo2022deepcore}, provide systematic tools for selecting representative subsets of training data in order to reduce computational cost while maintaining competitive performance.
In parallel, optimization-focused research has proposed adaptive optimizers and learning rate scheduling techniques, including AdaGrad~\cite{duchi2011adagrad}, RMSProp~\cite{tieleman2012rmsprop}, Adam~\cite{kingma2015adam}, and cyclical learning rates~\cite{smith2018disciplined}, as well as analyses of batch size effects on generalization~\cite{keskar2017largebatch}.
While these approaches have demonstrated notable improvements, most operate at the individual sample level, rely on predefined curricula, or modify the gradient update mechanism through reweighting or pruning. In contrast, the method proposed in this work introduces a lightweight A*-inspired heuristic that operates at the mini-batch level and frames batch scheduling as a heuristic search problem. Rather than pruning or reweighting individual samples, we prioritize entire batches based on a balanced estimate of difficulty and exploration, thereby preserving the original data distribution while guiding the training process toward more informative regions of the search space. This perspective bridges search-based decision strategies with stochastic optimization in deep learning, introducing a distinct direction compared to existing data-centric acceleration methods.

To make the boundary with prior work explicit, A*-BS differs from classical curriculum learning~\cite{bengio2009curriculum,kumar2010self} in two structural ways. First, classical curricula rely on an \emph{external}, predefined difficulty ordering (defined offline, e.g.\ by image complexity or a teacher model), whereas A*-BS uses a fully \emph{internal}, model-dependent difficulty signal which is recomputed at every epoch from the current network state. Second, curriculum learning typically presents easy samples first and hard samples later, whereas A*-BS prioritizes high-loss batches throughout training but actively counteracts the over-exploitation of any single batch through the reuse penalty $g(B_i)$. Compared with sample-level importance sampling~\cite{loshchilov2016online,katharopoulos2018importance}, A*-BS does not modify the gradient via per-sample weighting, so it preserves the original empirical distribution of each batch and is compatible with unmodified stochastic gradient descent; it only changes the \emph{order} in which batches are visited within an epoch.

\section{Methodology}
\subsection{Dataset Description}

To conduct the experiments of this study the MedMNIST\cite{medmnistv2} dataset was used as a benchmark. This dataset contains collections of medical images in standardized form. This in turn facilitates reproducible and fair comparisons between different machine learning models. Part of MedMNIST are multiple sub-datasets originating from different medical imaging modalities and clinical tasks. All of them are unified under a common data format and resolution.

\begin{figure*}[h!]
    \centering
    \includegraphics[width=\linewidth]{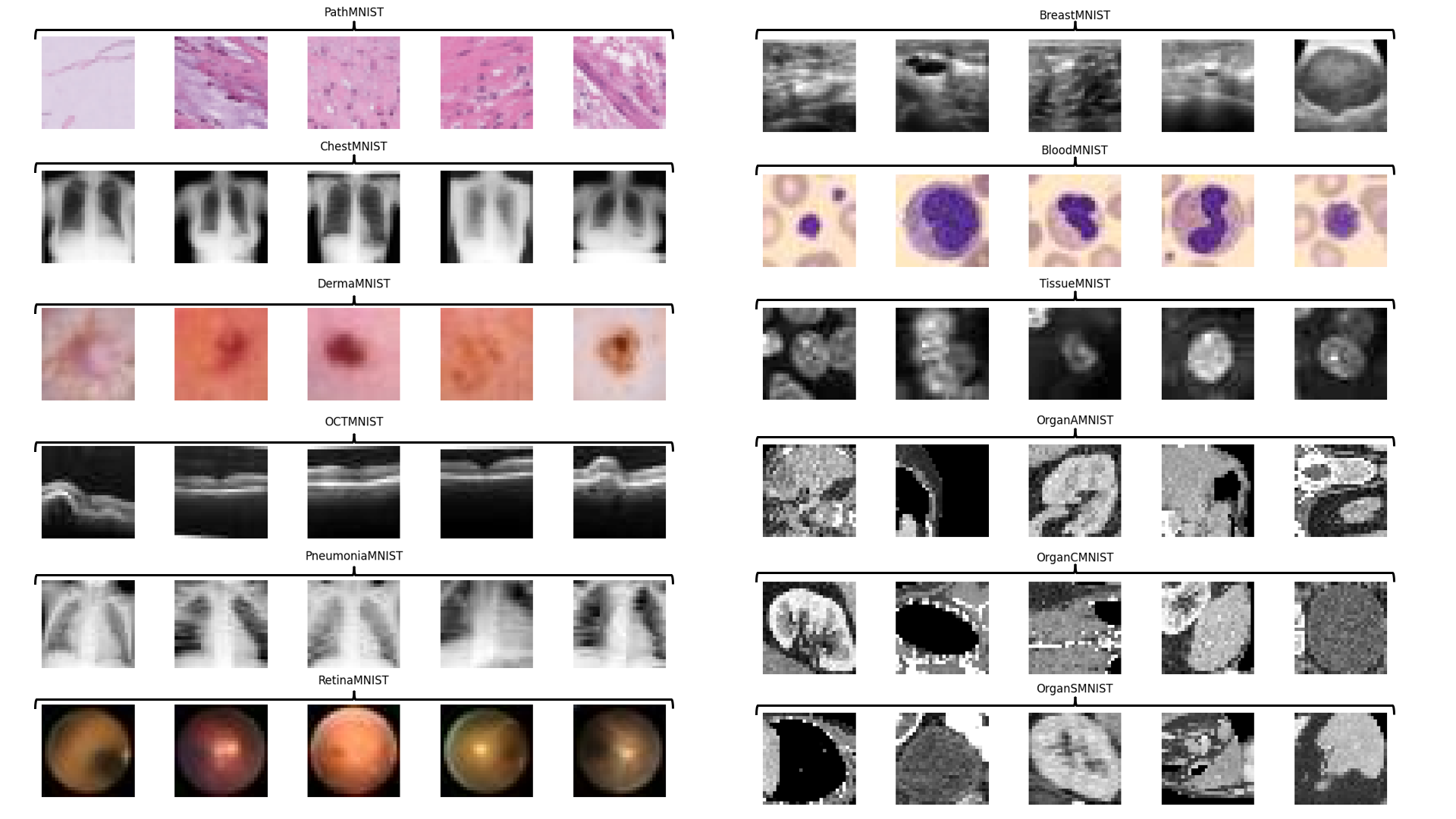}
    \caption{Representative visualization of the MedMNIST benchmark. For each sub-dataset, five sample images from the training set are presented, providing an overview of the heterogeneous imaging modalities, anatomical structures, and classification challenges addressed within the MedMNIST collection.}
    \label{subdatasets}
\end{figure*}

Although MedMNIST includes both 2D and 3D datasets, our work focuses exclusively on the 2D sub-datasets (or MedMNIST2D), as they are more appropriate for convolutional neural network (CNN) image classification. In particular, we consider the following 2D datasets: \textit{PathMNIST}, \textit{DermaMNIST}, \textit{ChestMNIST}, \textit{OCTMNIST}, \textit{PneumoniaMNIST}, \textit{BreastMNIST}, \textit{RetinaMNIST}, \textit{BloodMNIST}, \textit{TissueMNIST}, \textit{OrganAMNIST}, \textit{OrganSMNIST}, and \textit{OrganCMNIST} (See Fig. \ref{subdatasets}).

All datasets provided through the official MedMNIST package are standardized to a spatial resolution of $28 \times 28$ pixels. Depending on the specific sub-dataset, images are either grayscale (single-channel) or RGB (three-channel). In addition, the classification tasks vary across datasets, ranging from binary to multi-class problems. Each sub-dataset is supplied with predefined training, validation, and test splits, ensuring consistent evaluation protocols across experiments.

To account for these variations, the input and output dimensions of the proposed model are adapted automatically based on the number of image channels and the number of target classes associated with each sub-dataset and images are normalized to $[0,1]$.

\subsection{Model Architecture}

We used a lightweight CNN architecture tailored to the low spatial resolution of the MedMNIST images and intentionally kept the model simple, so that we could reduce overfitting and allow a clearer analysis of our proposed training strategy.

The network consists of two convolutional blocks followed by two fully connected layers, with the final layer dedicated to classification. Each convolutional block includes a $3 \times 3$ convolutional layer (with ReLU activation), followed by a $2 \times 2$ max-pooling operation for spatial downsampling. The number of feature maps increases across the convolutional layers, enabling the network to capture progressively more abstract visual patterns.

After the convolutional stages, the extracted feature maps are flattened and passed through a dense layer with ReLU activation. The final dense layer uses a softmax(or sigmoid) activation function based on the task, for multi-class or multi-label classification. The number of neurons in this layer is adjusted according to the number of classes of the corresponding sub-dataset.

The input shape of the network is adapted to the image type, using either $28 \times 28 \times 1$ for grayscale images or $28 \times 28 \times 3$ for RGB images. Overall, the architecture can be summarized as follows:

\begin{verbatim}
Input (28 x 28 x C)
-> Conv2D (32 filters, 3 x 3) + ReLU
-> MaxPooling (2 x 2)
-> Conv2D (64 filters, 3 x 3) + ReLU
-> MaxPooling (2 x 2)
-> Flatten
-> Dense (128) + ReLU
-> Dense (N_classes) + Softmax(or Sigmoid)
\end{verbatim}

The total number of trainable parameters of this architecture is approximately $2.25 \times 10^{5}$ (precisely, $224{,}002$ for the single-channel two-class case up to $225{,}481$ for the three-channel nine-class case), which remains roughly two orders of magnitude smaller than ResNet-18 ($\sim 1.12 \times 10^{7}$) and ResNet-50 ($\sim 2.36 \times 10^{7}$).

This architecture provides a good balance between simplicity and representational capacity, making it well suited for low-resolution medical image classification tasks while avoiding unnecessary architectural complexity.

\subsection{A*-Inspired Batch Selection}

The core idea is to treat each mini-batch as a ``node'' in a search space. At each epoch, batches are ranked using an A*-like score that combines a heuristic estimate $h(B_i)$ of how informative a batch is with a historical cost $g(B_i)$ that tracks how often each batch has already been used.

The heuristic term on multi-class classification tasks is defined as the average categorical cross-entropy computed from the model predictions:

\begin{equation}
h_c(B_i) = \frac{1}{N_i} \sum_{j=1}^{N_i} \text{CE}(y_j, \hat{y}_j),
\end{equation}

where $B_i$ is the $i$-th batch, $N_i$ is the number of samples in the batch, $y_j$ is the true label, $\hat{y}_j$ is the predicted label, and $\text{CE}$ is the categorical cross-entropy loss.

In multi-label classification tasks, such as ChestMNIST, each sample may belong to multiple classes simultaneously. In this setting, the categorical cross-entropy is replaced with binary cross-entropy applied independently to each label. Rather than computing a single global batch loss, we adopt a class-wise formulation where the heuristic is evaluated per label:

\begin{equation}
h_b(B_i) = \frac{1}{N_i} \sum_{j=1}^{N_i} \text{BCE}(y_{j,b}, \hat{y}_{j,b}),
\end{equation}

where $\text{BCE}$ denotes the binary cross-entropy for class $b$.

To incorporate exploration, we define the A*-inspired selection score:

\begin{equation}
f(B_i) = g(B_i) + \lambda \, h(B_i),
\end{equation}

where $\lambda$ controls the contribution of the heuristic term, and $h$ denotes the heuristic ($h_c$ for multi-class classification tasks, $h_b$ for multi-label classification tasks).

In contrast to standard A* formulations where $g(\cdot)$ represents a cumulative path cost, in our training framework $g(B_i)$ is incremented every time batch $B_i$ is selected for training. This makes previously used batches progressively less attractive even when their loss remains high, so the algorithm naturally balances exploitation of informative batches against exploration of underutilized ones.

\paragraph{Automatic computation of $\lambda$.}
Rather than fixing $\lambda$ manually, we compute it automatically at the beginning of each training epoch in order to balance the numerical scales of $g(B_i)$ and $h(B_i)$. Specifically, $\lambda$ is defined as the ratio between the standard deviations of the historical cost and the heuristic loss across mini-batches:

\begin{equation}
\lambda = \frac{\sigma_g}{\sigma_h + \varepsilon},
\end{equation}

where $\sigma_g$ and $\sigma_h$ denote the standard deviations of ${g(B_i)}$ and ${h(B_i)}$ over the set of batches respectively, and $\varepsilon$ is a small positive constant (set to $10^{-8}$ in our implementation) added to prevent division with zero and ensure numerical stability when $\sigma_h$ becomes very small.

This adaptive formulation ensures that neither the repetition penalty nor the loss-based heuristic dominates the selection process. Early in training, when batch reuse counts are similar, $\lambda$ remains small and the selection is primarily driven by the heuristic loss. As training progresses and batch reuse becomes more uneven, $\lambda$ increases accordingly, strengthening the exploration effect. As a result, the batch selection strategy remains stable throughout training while dynamically adapting to the evolving learning dynamics of the model.

\subsection{Implementation Details}

All experiments were performed in Google Colaboratory (Colab) using Python 3.12.12 with GPU acceleration enabled. Model training was conducted on an NVIDIA Tesla T4 GPU provided by the Colab runtime environment.

The software stack included NumPy 2.0.2, TensorFlow 2.19.0 (with Keras 3.10.0), Scikit-learn 1.6.1, and MedMNIST 3.0.2.

All experiments were executed under the default Colab runtime configuration available at the time of experimentation.

For every dataset, the lightweight CNN was trained for the same number of epochs and with the same Adam optimizer and learning rate under both the random-shuffling baseline and the A*-BS strategy, so that the only varying factor is the batch-ordering policy. The heuristic in Eq.~(1)/(2) is computed using a stochastic subsample of $32$ examples per batch, which keeps the per-epoch overhead of the ranking step approximately linear in the number of batches; the actual measured overhead is reported in Sec.~IV-C. 

\begin{table*}[!h]
\centering
\caption{Performance comparison (ACC and AUC) of A*-BS used in a simple CNN model with ResNet-18 and ResNet-50 on MedMNIST datasets.}
\label{tab:medmnist_results}
\renewcommand{\arraystretch}{1.2}
\begin{tabular}{llcccccc}
\hline
Nr: & \textbf{Dataset}
& \multicolumn{2}{c}{\textbf{A*-BS}}
& \multicolumn{2}{c}{\textbf{ResNet-18\cite{medmnistv2}}}
& \multicolumn{2}{c}{\textbf{ResNet-50\cite{medmnistv2}}} \\
\cline{2-7}
& & ACC & AUC & ACC & AUC & ACC & AUC \\
\hline
1 & PathMNIST        & 0.835 & 0.984 & 0.907 & 0.983 & \textbf{0.911} & \textbf{0.990} \\
2 & ChestMNIST       & 0.6397 & 0.746 & 0.947 & 0.768 & 0.947 & 0.769 \\
3 & DermaMNIST       & \textbf{0.753} & 0.908 & 0.735 & \textbf{0.917} & 0.735 & 0.913 \\
4 & OCTMNIST         & \textbf{0.897} & \textbf{0.978} & 0.743 & 0.943 & 0.762 & 0.952 \\
5 & PneumoniaMNIST   & \textbf{0.958} & \textbf{0.996} & 0.854 & 0.944 & 0.854 & 0.948 \\
6 & BreastMNIST      & \textbf{0.897} & 0.893 & 0.863 & \textbf{0.901} & 0.812 & 0.857 \\
7 & RetinaMNIST      & \textbf{0.533} & \textbf{0.803} & 0.524 & 0.717 & 0.528 & 0.726 \\
8 & TissueMNIST      & 0.616 & 0.895 & 0.676 & 0.930 & \textbf{0.680} & \textbf{0.931} \\
9 & BloodMNIST       & 0.938 & 0.996 & \textbf{0.958} & \textbf{0.998} & 0.956 & 0.997 \\
10 & OrganAMNIST     & \textbf{0.955} & \textbf{0.999} & 0.935 & 0.997 & 0.935 & 0.997 \\
11 & OrganSMNIST     & \textbf{0.852} & \textbf{0.989} & 0.782 & 0.972 & 0.770 & 0.972 \\
12 & OrganCMNIST     & \textbf{0.956} & \textbf{0.999} & 0.900 & 0.992 & 0.905 & 0.992 \\
\hline
\end{tabular}
\end{table*}

\section{RESULTS}

After constructing the proposed model using the A*-inspired batch selection (A*-BS) strategy, which prioritizes training samples from more difficult to easier ones, we conducted a comparative evaluation against two widely used convolutional neural network architectures, namely ResNet-18 and ResNet-50. In order to ensure a fair and consistent comparison between the models chosen for this study, all of them were evaluated on the same classification tasks and experimental settings reported in the MedMNIST-v2 benchmark~\cite{medmnistv2}.

The evaluation focuses on classification tasks involving $28 \times 28$ images provided by the MedMNIST-v2 2D datasets, which are also employed in this study. As for performance, it was assessed through the use of two standard metrics: classification accuracy (ACC) and area under the ROC curve (AUC). After completing the twelve classification tasks we have summarized the results in Table~\ref{tab:medmnist_results}.

\subsection{Scope of the comparison with ResNet}
We emphasize that the ResNet-18 and ResNet-50 results displayed in Table~\ref{tab:medmnist_results} are the values originally reported in the MedMNIST-v2 reference paper~\cite{medmnistv2}, and were not re-trained in this study. They follow the training protocol described in~\cite{medmnistv2}, which is publicly documented and identical across all twelve MedMNIST 2D tasks: 100 training epochs, cross-entropy loss (or binary cross-entropy for multi-label tasks), and Adam optimizer with the schedule reported by~\cite{medmnistv2}. We deliberately do not re-train ResNet-18/-50 from scratch in our own pipeline, in order to (i) avoid creating an unintended advantage by tuning the baselines under our own infrastructure, and (ii) compare A*-BS directly against the publicly accepted reference numbers for MedMNIST-v2. Consequently, Table~\ref{tab:medmnist_results} should be interpreted as a comparison of the lightweight-CNN+A*-BS pipeline with publicly reported benchmark figures rather than as a controlled head-to-head experiment under identical hyper-parameters, augmentation, and training time. The ablation reported in Sec.~IV-B and the wall-clock comparison reported in Sec.~IV-C address the controlled-comparison concern by varying only the batch-ordering policy on a single fixed architecture, and by reporting per-epoch training cost on identical hardware, respectively.

From the table above, it can be observed that the proposed A*-BS strategy achieves superior performance in six out of the twelve classification tasks when compared to both ResNet-18 and ResNet-50, considering both ACC and AUC metrics. The relative improvements range from approximately 1\% to 15\%, depending on the dataset and task complexity. In two additional classification tasks, the proposed approach achieves improved performance in terms of classification accuracy, while remaining competitive in AUC.

We also explicitly acknowledge that on several datasets, namely ChestMNIST, TissueMNIST and BloodMNIST, ResNet-18 / ResNet-50 outperform the lightweight CNN combined with A*-BS, both in ACC and AUC. This is consistent with the expected behavior of a model with $\sim\!2 \times 10^{5}$ parameters on tasks where additional representational depth is genuinely useful.

A broader comparison with previously reported results on MedMNIST-2D, including ResNet-18, ResNet-50\cite{resnet} trained on $28 \times 28$ images, ResNet variants trained on higher-resolution inputs ($224 \times 224$), as well as automated machine learning approaches such as auto-sklearn, AutoKeras, and Google AutoML Vision, further demonstrates the effectiveness of the proposed approach. In particular, the A*-BS strategy combined with a lightweight CNN achieves competitive or superior performance in both ACC and AUC across six classification tasks, namely OCTMNIST, PneumoniaMNIST, RetinaMNIST, OrganAMNIST, OrganCMNIST, and OrganSMNIST.

\subsection{Ablation: same lightweight CNN with vs.\ without A*-BS}

To isolate the contribution of the proposed strategy from the contribution of the underlying architecture, we additionally trained the same lightweight CNN (Sec.~III-B) on every MedMNIST 2D task under two configurations that differ \emph{only} in the batch-ordering policy: standard random shuffling, and A*-BS. Both configurations use identical architecture, optimizer, learning rate, batch size, hardware, and number of training epochs. For each (dataset, strategy) pair, the test ACC and AUC reported in Table~\ref{tab:ablation} correspond to the best validation-ACC checkpoint of a single training run. Across the twelve datasets, A*-BS improves test ACC and AUC over the random-shuffling baseline on all datasets (Table~\ref{tab:medmnist_results}).
\begin{table}[!h]
\centering
\caption{Ablation: lightweight CNN trained with random shuffling vs.\ A*-BS, same architecture and hyper-parameters, single training run per cell. Bold marks the better of the two strategies per metric.}
\label{tab:ablation}
\renewcommand{\arraystretch}{1.15}
\begin{tabular}{lcccc}
\hline
\textbf{Dataset} & \multicolumn{2}{c}{\textbf{CNN + random}} & \multicolumn{2}{c}{\textbf{CNN + A*-BS}} \\
\cline{2-5}
                 & ACC & AUC & ACC & AUC \\
\hline
PathMNIST        & 0.745 & 0.951 & \textbf{0.835} & \textbf{0.984} \\
ChestMNIST & 0.199 & 0.724 & \textbf{0.640} & \textbf{0.746} \\
DermaMNIST       & 0.734 & 0.897 & \textbf{0.753} & \textbf{0.908} \\
OCTMNIST         & 0.716 & 0.925 & \textbf{0.897} & \textbf{0.978} \\
PneumoniaMNIST   & 0.859 & 0.950 & \textbf{0.958} & \textbf{0.996} \\
BreastMNIST      & 0.827 & 0.854 & \textbf{0.897} & \textbf{0.893} \\
RetinaMNIST      & 0.513 & 0.734 & \textbf{0.533} & \textbf{0.803} \\
TissueMNIST      & 0.599 & 0.888 & \textbf{0.616} & \textbf{0.895} \\
BloodMNIST       & 0.893 & 0.989 & \textbf{0.938} & \textbf{0.996} \\
OrganAMNIST      & 0.820 & 0.960 & \textbf{0.955} & \textbf{0.999} \\
OrganSMNIST      & 0.656 & 0.939 & \textbf{0.852} & \textbf{0.989} \\
OrganCMNIST      & 0.7978   & 0.968    & \textbf{0.956} & \textbf{0.999} \\
\hline

\end{tabular}
\end{table}

\subsection{Training-time and computational-cost analysis}

We additionally evaluate the computational cost of the proposed A*-BS strategy by measuring the wall-clock training time of the lightweight CNN with random shuffling, the same CNN with A*-BS, and the ResNet-18 and ResNet-50 baselines. All models were trained from scratch on the same NVIDIA Tesla T4 GPU for 20 epochs using identical optimizer settings and batch size ($128$).

The additional cost of A*-BS consists of a lightweight forward pass on a 32-sample subset per batch to estimate $h(B_i)$, followed by batch sorting once per epoch. Table~\ref{tab:timing} reports the mean per-epoch and total training time on four MedMNIST datasets.

\begin{table}[h]
\centering
\caption{Per-epoch and 20-epoch wall-clock training time on a single NVIDIA Tesla T4 (Google Colab). Same batch size ($128$), same optimizer, identical hardware, on three different datasets.}
\label{tab:timing}
\renewcommand{\arraystretch}{1.15}
\begin{tabular}{p{2.5cm} p{1.2cm} p{1.4cm} p{1.6cm}}
\hline
\textbf{Model / strategy} & \textbf{Params} & \textbf{s / epoch} & \textbf{Total (20 ep.) [s]} \\
\hline
\multicolumn{4}{l}{\emph{PathMNIST} (89{,}996 train, 9 classes, RGB)} \\
\hline
CNN + random   & $225{,}481$    &  41.26 &    825.2 \\
CNN + A*-BS    & $225{,}481$    &  50.19 &  1{,}003.7 \\
ResNet-18      & $11{,}183{,}049$ & 362.50 &  7{,}250.0 \\
ResNet-50      & $23{,}571{,}913$ & $980.4$ & 19{,}608 \\
\hline
\multicolumn{4}{l}{\emph{RetinaMNIST} (1{,}080 train, 5 classes, RGB)} \\
\hline
CNN + random   & $224{,}965$    &   2.64 &     52.75 \\
CNN + A*-BS    & $224{,}965$    &   1.95 &     39.04 \\
ResNet-18      & $11{,}180{,}997$ &  89.34 &  1{,}786.85 \\
ResNet-50      & $23{,}563{,}717$ & 249.45 &  4{,}988.93 \\
\hline
\multicolumn{4}{l}{\emph{BreastMNIST} (546 train, 2 classes, grayscale)} \\
\hline
CNN + random   & $224{,}002$    &   0.71 &     14.29 \\
CNN + A*-BS    & $224{,}002$    &   0.87 &     17.44 \\
ResNet-18      & $11{,}178{,}306$ &  43.20 &    863.92 \\
ResNet-50      & $23{,}556{,}418$ & 121.22 &  2{,}424.40 \\
\hline
\multicolumn{4}{l}{\emph{PneumoniaMNIST} (4{,}708 train, 2 classes, grayscale)} \\
\hline
CNN + random   & $224{,}002$    &   2.20 &     43.94 \\
CNN + A*-BS    & $224{,}002$    &   2.98 &     59.57 \\
ResNet-18      & $11{,}178{,}306$ &  19.37 &    387.47 \\
ResNet-50      & $23{,}556{,}418$ &  52.26 &  1{,}045.14 \\
\hline
\end{tabular}
\end{table}

The results show that A*-BS introduces only a modest overhead compared with random shuffling, while remaining substantially more computationally efficient than ResNet-18 and ResNet-50. Moreover, the lightweight CNN combined with A*-BS achieves competitive performance within the same 20-epoch training regime, indicating improved convergence efficiency.

\section{CONCLUSIONS}
In this work, we introduced an A*-inspired batch selection (A*-BS) strategy designed to improve the training efficiency and convergence behavior of convolutional neural networks without modifying their architecture or optimization algorithms. By formulating the mini-batch scheduling process as a heuristic-driven search problem, the proposed method dynamically prioritizes informative batches while discouraging excessive reuse, thereby balancing exploitation and exploration during training.

The proposed approach was evaluated across diverse imaging modalities, varying class distributions, and different levels of task difficulty using a lightweight CNN architecture in order to demonstrate its competitive performance and, in several cases, its superiority compared to substantially deeper and more complex models.

As demonstrated in Table~\ref{tab:medmnist_results}, the proposed method outperformed ResNet-18 and ResNet-50 in terms of both accuracy and AUC in six out of twelve classification tasks, with relative improvements ranging from approximately 1\% to 15\%. In additional tasks, comparable or improved classification accuracy was observed, highlighting the robustness of the proposed strategy across heterogeneous datasets. These results indicate that intelligent batch selection can compensate for reduced architectural complexity. On the controlled comparison, A*-BS improves ACC and AUC over random shuffling on all the twelve datasets evaluated. On the datasets for which ResNet retains an advantage in Table~\ref{tab:medmnist_results} (notably ChestMNIST, TissueMNIST and BloodMNIST), A*-BS does not close the gap, and we therefore avoid presenting it as a replacement for deeper architectures.
But it can also be used in deeper CNN architectures, and referring to Table II, its integration into CNN shows a significant improvement when using A*-BS compared to when we do not use it.

\begin{table}[t]
\centering
\caption{Comparison of the proposed simple CNN Architecture with batch selection with ResNet-Based Models used in the some classification tasks.}
\label{tab:architecture_comparison}
\renewcommand{\arraystretch}{1.15}
\begin{tabular}{p{1.4cm} p{1.0cm} p{1.4cm} p{3.0cm}}
\hline
\textbf{Model} & \textbf{Depth} & \textbf{Parameters} & \textbf{Key Feature} \\
\hline
Baseline CNN & 6 &  $\sim 2.25 \times 10^{5}$ & Linear convolutional blocks \\
ResNet-18 & 18 & $1.12 \times 10^{7}$ & Residual blocks with skip connections \\
ResNet-50 & 50 & $2.36 \times 10^{7}$ & Bottleneck residual blocks \\
\hline
\end{tabular}
\end{table}

Furthermore, the architectural comparison presented in Table~\ref{tab:architecture_comparison} highlights the significant difference in model complexity. While ResNet-based architectures rely on deep residual and bottleneck blocks with millions of parameters, the proposed CNN contains on the order of $2 \times 10^{5}$ parameters. Nevertheless, when combined with the A*-BS strategy, the lightweight model demonstrates strong generalization performance in more than 70\% of the evaluated tasks, confirming that training dynamics play a critical role alongside model capacity.

Overall, the results suggest that the proposed A*-inspired batch selection strategy is a useful complementary training-time mechanism whose computational overhead is bounded and small (Sec.~IV-C, approximately $22{-}35\%$ over random shuffling on the lightweight CNN), and which can improve the training dynamics on all of MedMNIST-v2 tasks (Sec.~IV-B). 

\subsection*{Limitations and future work}
The empirical results reported here have three identified limitations. First, the comparison against ResNet-18 / ResNet-50 in Table~\ref{tab:medmnist_results} uses the official MedMNIST-v2 results and is therefore not a controlled head-to-head experiment under identical infrastructure; the controlled element of the comparison is provided by the ablation in Sec.~IV-B and by the wall-clock comparison in Sec.~IV-C. Second, the ablation in Table~\ref{tab:ablation} reports a single training run per (dataset, strategy) pair; a multi-seed expansion ($R \geq 5$ seeds per cell) and a paired statistical test between the two strategies are identified as the next experimental step. Third, the gains of A*-BS are dataset-dependent and not universal, and the present analysis is restricted to 2D, low-resolution, single-modality medical imaging at $28 \times 28$. Higher-resolution, 3D, and multi-modal data are left for future work, along with the combination of A*-BS with adaptive optimizers.

\end{document}